%% file: main.tex
\documentclass{article}
\pdfoutput=1
\PassOptionsToPackage{numbers, compress}{natbib}
\usepackage[final]{nips_2017}

\usepackage[utf8]{inputenc} % allow utf-8 input
\usepackage[T1]{fontenc}    % use 8-bit T1 fonts
\usepackage{hyperref}       % hyperlinks
\usepackage{url}            % simple URL typesetting
\usepackage{booktabs}       % professional-quality tables
\usepackage{amsfonts}       % blackboard math symbols
\usepackage{nicefrac}       % compact symbols for 1/2, etc.
\usepackage{microtype}      % microtypography
\usepackage{amsmath}
\usepackage{amssymb}
\usepackage{graphicx}
\usepackage{color}
\usepackage{float}
\usepackage{wrapfig}
\usepackage{epigraph}

\makeatletter
\@ifpackagewith{nips_2017}{final}{
\newcommand{\note}[1]{}
\newcommand{\todo}[1]{}
\newcommand{\hideurl}[1]{\url{#1}}
}{
\newcommand{\note}[1]{{\color{red}\textbf{#1}}}
\newcommand{\todo}[1]{\note{TODO: #1}}
\newcommand{\hideurl}[1]{\url{https://youtu.be/videos.for.reviewers.are.on.cmt}}
}
\makeatother

\newcommand{\T}{\mathrm{T}}
\newcommand{\AND}{\operatorname{and}}
\newcommand{\OR}{\operatorname{or}}
\newcommand{\NOT}{\operatorname{not}}
\newcommand{\vPhi}{\boldsymbol{\Phi}}
\newcommand{\vO}{\boldsymbol{\Omega}}

\title{Programmable Agents}

\author{
  Misha Denil \quad Sergio G\'omez Colmenarejo \\ \textbf{Serkan Cabi \quad David Saxton \quad Nando de Freitas} \\
  DeepMind \\
  \texttt{\{mdenil,sergomez,cabi,saxton,nandodefreitas\}@google.com}
}

\begin{document}
\maketitle

\begin{abstract}
  We build deep RL agents that execute declarative programs expressed in formal language.  The agents learn to ground the terms in this language in their environment, and can generalize their behavior at test time to execute new programs that refer to objects that were not referenced during training.  The agents develop disentangled interpretable representations that allow them to generalize to a wide variety of zero-shot semantic tasks.
\end{abstract}

\input{010-introduction}

% The NIPS footnote on the first page is a bottom float, the introduction must be included before setting this counter.
\setcounter{bottomnumber}{0}

\input{020-related-work}
\input{030-tasks-as-programs}
\input{040-programmable-environments}
\input{050-programmable-networks}
\input{060-experiments}
\input{070-conclusion}

\clearpage

{\small
\bibliographystyle{abbrv}
\bibliography{programmable-agents}
}

\clearpage

\appendix

\input{080-training-details}
\input{090-novel-objects}

\clearpage

\input{110-detector-visualizations}

\end{document}

%% file: 010-introduction.tex
\section{Introduction}

This paper shows how to build agents that can execute declarative programs expressed in a simple formal language.  The agents learn to ground the terms of the language in their environment through experience.  The learned groundings are disentangled and compositional; at test time we can ask the agents to perform tasks that involve novel combinations of properties and they will do so successfully.

The agents learn to distinguish distinct properties that are referenced together during training; when trained on tasks that always reference objects through a conjunction of shape and color the agents can generalize at test time to tasks that reference objects through either property in isolation.

Completely novel object properties can be referenced through the principle of exclusion (i.e.\ the object whose color you have not seen before), and our agents are able to successfully complete tasks that reference novel objects in this way.  This works even when the agents have never encountered programs involving this type of reference during training. %Agents also succeed in referring to objects that possess multiple novel properties, and via combinations of known and unknown properties.
Referring to objects that possess multiple novel properties is also successful, as is referring to objects through combinations of known and unknown properties.

Our agents are robust to catastrophic forgetting.  If we train on a subset of tasks then not only can the agents accomplish novel tasks zero-shot, but if we switch to training only on the novel tasks then performance on the original tasks does not degrade.

Our agents are implemented as deep neural networks, and trained end to end with reinforcement learning.  The agents learn how programs refer to properties of objects and how properties are assigned to objects in the world entirely through their experience interacting with their environment.  Natural and interpretable assignments of properties to objects emerge without any direct supervision of how these properties should be assigned.  No auxiliary prediction or control tasks are required for good performance (although exploring their effects in this setting is a nice possibility for future work).

The representations learned by our agents are extremely interpretable.  There is an explicit mapping between activation maps in the programmable layers of our agents and terms in the programs they execute.  Visualizing these activation maps provides an immediately interpretable picture of how the agents assign properties to objects in their world.

The extremely powerful generalization our agents achieve is made possible by our novel ``Programmable Network'' architecture.  Programmable Networks impose a sophisticated bottleneck on the agent's representations whose structure ensures that their representations will generalize.

% (semantic) diversity of tasks 
% [To match the combinatorial complexity of the environment and succeed, we need a combinatorial agent]

%% file: 020-related-work.tex
\section{Related work}

The work in this paper builds on two recent lines of work in deep learning.  One is the resurgence of modular networks that can be decomposed and recomposed to compute different functions.  Neural Module Networks~\cite{Andreas2016a, Andreas2016b}, which use modularity in visual question answering, have been particularly influential on our thinking.  This type of approach has also recently seen great success in answering questions that involve relational reasoning~\cite{Hu2017, johnson2017}. 

The second recent trend we follow is the rise of relational neural network architectures, particularly Interaction Networks~\cite{Battaglia2016a} and the Neural Physics Engine~\cite{chang2017}.  At a high level these works on relational architectures are subsumed by the various neural approaches to graph processing.  A nice overview of recent techniques can be found in Gilmer~et~al.~\cite{gilmer2017neural}.

Compositionality has become a topic of great interest in machine learning, robotics and cognitive science \cite{Lake2017,reed2016,kulkarni2016hierarchical,devin2016learning,tran2017deep,misra2017composing}, and it has been central to research in language \cite{socher2012semantic,mikolov2013distributed,socher2013recursive,yogatama2016learning}.
Compositional neural architectures paired with different forms of attention have led to impressive results in natural language interfaces for database tables, and language acquisition in 2D navigation environments \cite{Neelakantan2017learning,Yu2017}.  Yu~et~al.~\cite{Yu2017} present a different deep RL approach to grounding symbols on perception, and introduce a visual question answering auxiliary task to improve zero-shot generalization.
Use of programs to specify hard deterministic gating is also reminiscent of PAQ8~\cite{mahoney2005,knoll2012}.

% The SCONE dataset (\url{https://nlp.stanford.edu/projects/scone/}).  The main citation for SCONE is \cite{Long2016}, but there are other citations on the website as well.  SCONE is triples of initial state, language instruction, final state.  The task is to map from the language instruction to a logical program that produces the final state when executed on the initial state.  One of the goals of SCONE is to express different linguistic phenomena like anaphora.  The focus in SCONE is really on inferring the logical form, which is different than our goal.  This paper is really about a hierarchy of semantic parsing models.

% This paper about mapping natural language to programs of environment manipulation \cite{Guu2017}. This paper uses SCONE, it is more recent than the original SCONE paper.  It focuses on learning semantic parsing.  Sequence to sequence model with attention.  Something about comparing RL with MML (maximum marginal likelihood) for their objective.  Randomized beam search for avoiding ``spurious'' programs.

%% file: 030-tasks-as-programs.tex
\section{Tasks as declarative programs}

Building neural networks that infer or execute computer programs has been a popular topic of recent research.  Different approaches to this problem focus on networks that infer programs \cite{balog2017deepcoder,devlin2017robustfill,ling2017learning}, on networks that execute programs \cite{zaremba2014learning, reed2016, bosnjak2016} and on networks that jointly do both~\cite{grefenstette2015learning,kaiser2015neural, zaremba2016learning, Neelakantan2016neural, graves2016dnc}; although there is much overlap between many of these approaches.  The majority of these works model their networks after an imperative programming style, although researchers have also begun to explore structuring their networks as functional (recursive) programs as well~\cite{cai2017,reed2016}.

In this work we depart from previous work by considering networks that execute a simple \emph{declarative} language.  Paradigmatic examples of declarative languages are PROLOG~\cite{bratko2001prolog, bosnjak2016} and SQL.  
%Although the declarative paradigm has fallen out of favor in the AI community, it
The declarative paradigm
provides an appealing and flexible way to describe tasks for agents~\cite{Littman2017a}.

% The key difference between declarative programs and other approaches is that a declarative programs specifies \emph{what} must be done, but not \emph{how} the goal should be accomplished.  Taking this view allows us to specify desirable states of the world (goals) for our agents while allowing the agents to fill in the specific details of how these goals should be achieved.

Our general framework is as follows:  A \emph{goal} is specified as a state of the world that satisfies a \emph{relation} between two \emph{objects}.  Objects are associated with sets of \emph{properties} (e.g.\ their color and shape).  The vocabulary of properties gives rise to a system of \emph{base sets} which are the sets of objects that share each named property (e.g.\ \texttt{RED} is the set of red objects, etc).  The full universe of discourse is then the Boolean algebra generated by these base sets.

We require two things for each program.  The \emph{verifier} has access to the true state of the environment, and can inspect this state to determine if it satisfies the program.  We also need a \emph{search procedure} which inspects the program as well as some summary of the environment state and decides how to modify the environment to bring the program closer to satisfaction.

These components correspond directly to components of the standard RL setup.  Notably, the verifier is a reward function (which has access to privileged information about the environment state) and the search procedure is an agent (which may have a more restrictive observation space).

There are several advantages to thinking in this way.  The first is that building semantic tasks becomes straightforward, we need only specify a new program to obtain a new reward function that depends on semantic properties of objects in the environment. Consequently, %our agents are capable of solving
we can easily specifiy
diverse, combinatorial tasks.

Another advantage is that this framing places the emphasis squarely on generalization to new tasks.  A program interpreter is not very useful if you must enumerate all programs you might want to run up front.  Our goal is not only to perform combinatorial tasks, but to be able to specify new behaviors at test time, and for them to be accomplished successfully without additional training.

This type of generalization is quite difficult to achieve with deep RL.  In this paper we focus on reaching for blocks, but the simplicity of the individual tasks belies the complexity of what we achieve.  The challenge here is to have agents ground the terms of the programming language in their environment and, at test time, to be able to execute \emph{new} programs that use these terms in novel ways.

%% file: 040-programmable-environments.tex
\section{Programmable reaching environment}
\label{sec:environment}

\begin{figure}
\centering
\includegraphics[width=0.24\linewidth]{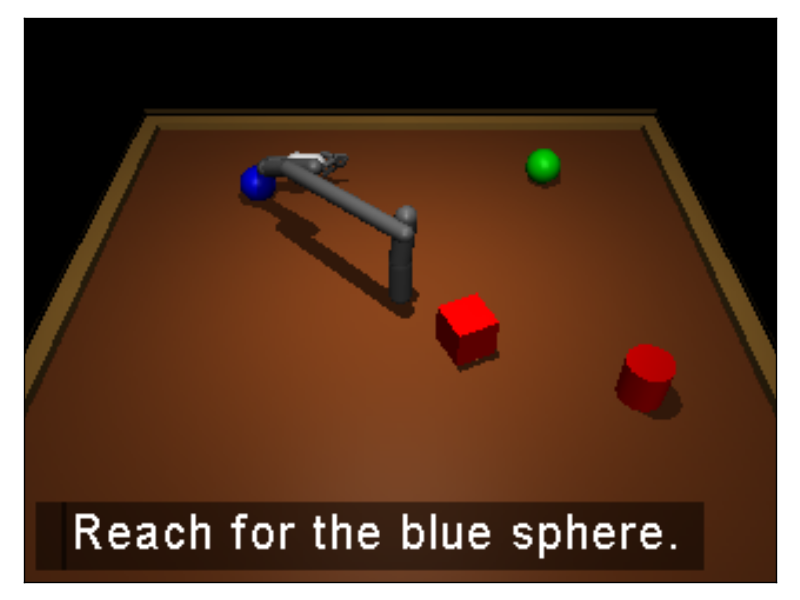}
\includegraphics[width=0.24\linewidth]{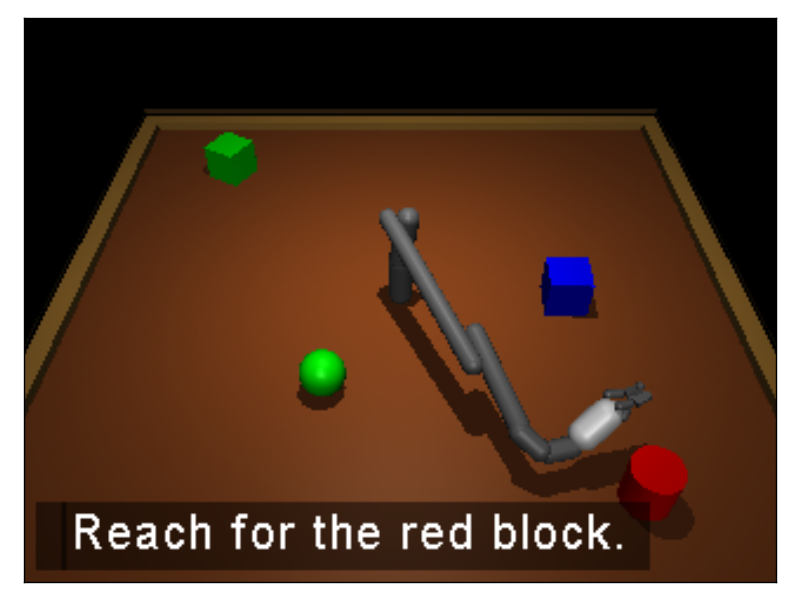}
\includegraphics[width=0.24\linewidth]{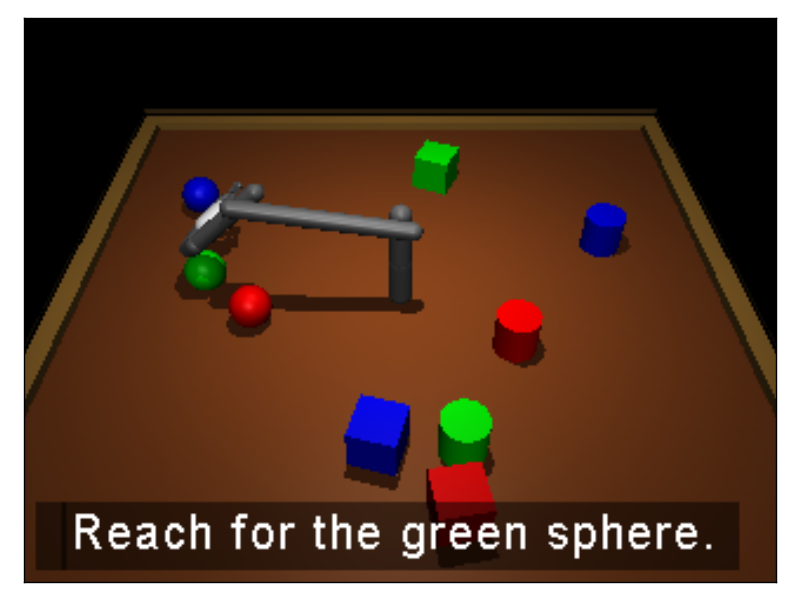}
\includegraphics[width=0.24\linewidth]{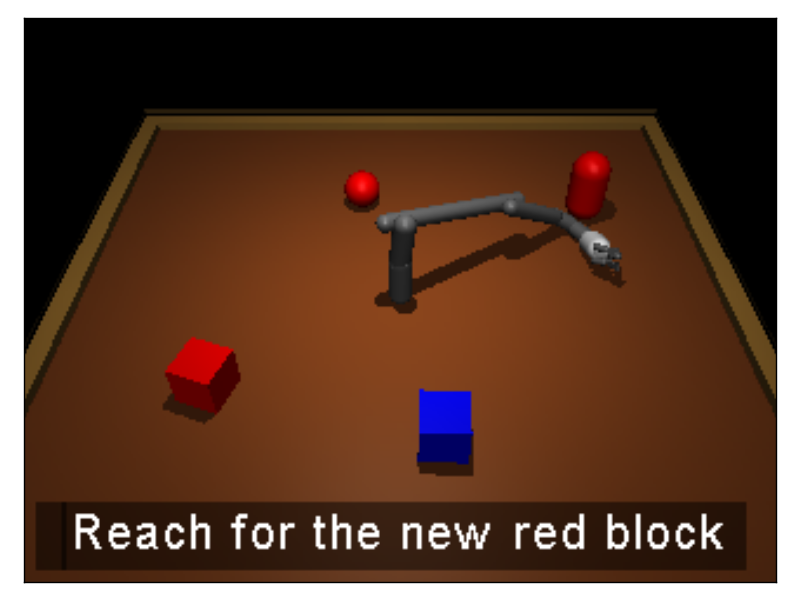}
\caption{The programmable reaching environment consists of a square arena with a robot arm in the center.  Blocks are scattered randomly and the task in each episode is to reach towards a specific block, identified by a combination of shape and color.  The leftmost frame shows a training episode, and the remaining frames show various types of generalization that agents trained in this setting are capable of, including operating on single properties, varying numbers of objects, and novel shapes.}
\label{fig:jaco-tabletop-reaching}
\end{figure}

Figure~\ref{fig:jaco-tabletop-reaching} shows several visualizations of our programmable reaching environment, which consists of a mechanical arm in the center of a large table.  The arm is a simplified version of the Jaco arm, where the body has been stereotyped to basic geoms (rigid body building components~\cite{todorov2012}), and the finger actuators have been disabled.  In each episode a fixed number of blocks appear at random locations on the table.  Each block has both a shape and a color, and the combination of both are guaranteed to uniquely identify each block within the episode.  The programmable reaching environment is implemented with the MuJoCo~\cite{todorov2012} physics engine, and hence the objects are subject to friction, contact forces, gravity, etc.

% \begin{figure}
%     \centering
%     \includegraphics[width=0.25\linewidth]{artwork/designs}
%     \caption{Different designs we use in our experiments.  Rows and columns of each matrix correspond to factors of variation, indexed by the levels they can take (we consider only two factor experiments in this paper).  Each cell of the matrix corresponds to a different task.  The cells of each matrix are color coded so that green indicates train conditions, and red indicates conditions that are not seen at train time, and are used to evaluate zero-shot generalization.}
%     \label{fig:designs}
% \end{figure}

Each task in the reaching environment is to put the ``hand'' of the arm (the large white geom) near the target block, which changes in each episode.  The task is communicated to the agent with two integers specifying the target color and shape, respectively.

The complexity of the environment can be varied by changing the number, colors and shapes that blocks can take.  In this work we consider 2x2 (two colors and two shapes) and 3x3 variants.  We can also control the number of blocks that appear on the table in each episode, and we fix it to four blocks during training to study generalization to other numbers.  When there are more possible blocks than are allowed on the table the episode generator ensures that the reaching task is always achievable (i.e.\ the agent is never asked to reach for a block that is not present).

The arm has 6 actuated rotating joints, which results in 6 continuous actions in the range $[0, 1]$.  The observable features of the arm are the positions of the 6 joints, along with their angular velocities.  The joint positions are represented as the $\sin$ and $\cos$ of the angle of the joint in joint coordinates.  This results in a total of 18 ($6 \times 2 + 6$) body features describing the state of the arm.

Objects are represented using their 3d position as well as a 4d quaternion representing their orientation, both represented in the coordinate frame of the hand.  Each block also has a 1-hot encoding of its shape (4d) and its color (5d), for a total of 16 object features per block.  We provide object features for all of the blocks on the table as well as the hand, but not for the other bodies that compose the arm.

We can write a reaching program in this environment as
\begin{align}
    \texttt{NEAR(HAND, AND(RED, CUBE))}
    \label{eq:program-example}
\end{align}
which specifies that the hand should be near the red cube.  This condition can be checked automatically by the verifier to produce a reward function that takes on the values $\pm 1$.  The verifier interprets \texttt{NEAR} by thresholding the distance between its arguments, but this is exposed to the agent only through the values the reward function takes in different states.

To show zero-shot generalization we partition the set of possible target blocks into train and test conditions.  We train an agent by choosing a target randomly from the train conditions in each episode, and evaluate the agent on its performance reaching for blocks from the test conditions, which were never targets during training.  We call an assignment of targets to train and test conditions a \emph{design}.

The left panel in Figure~\ref{fig:program-example} shows the designs we consider for the 2x2 and 3x3 variants of the reaching environment.  Rows and columns of each matrix correspond to different shapes and colors, respectively and each cell of each matrix corresponds to a different task.  The cells are color coded so that yellow indicates train conditions, and magenta indicates conditions only seen at test time.  The magenta tasks are used to evaluate zero-shot generalization after the agent is trained.

%% file: 050-programmable-networks.tex
\section{Programmable networks}

Due to space constraints we describe only the main components of the Programmable Network architecture in this section.  Full details of how everything comes together can be found in Appendix~\ref{sec:training-details}.

\subsection{Executing programs}
\label{sec:executing-programs}

\begin{figure}[t]
    \centering
    \includegraphics[width=0.155\linewidth]{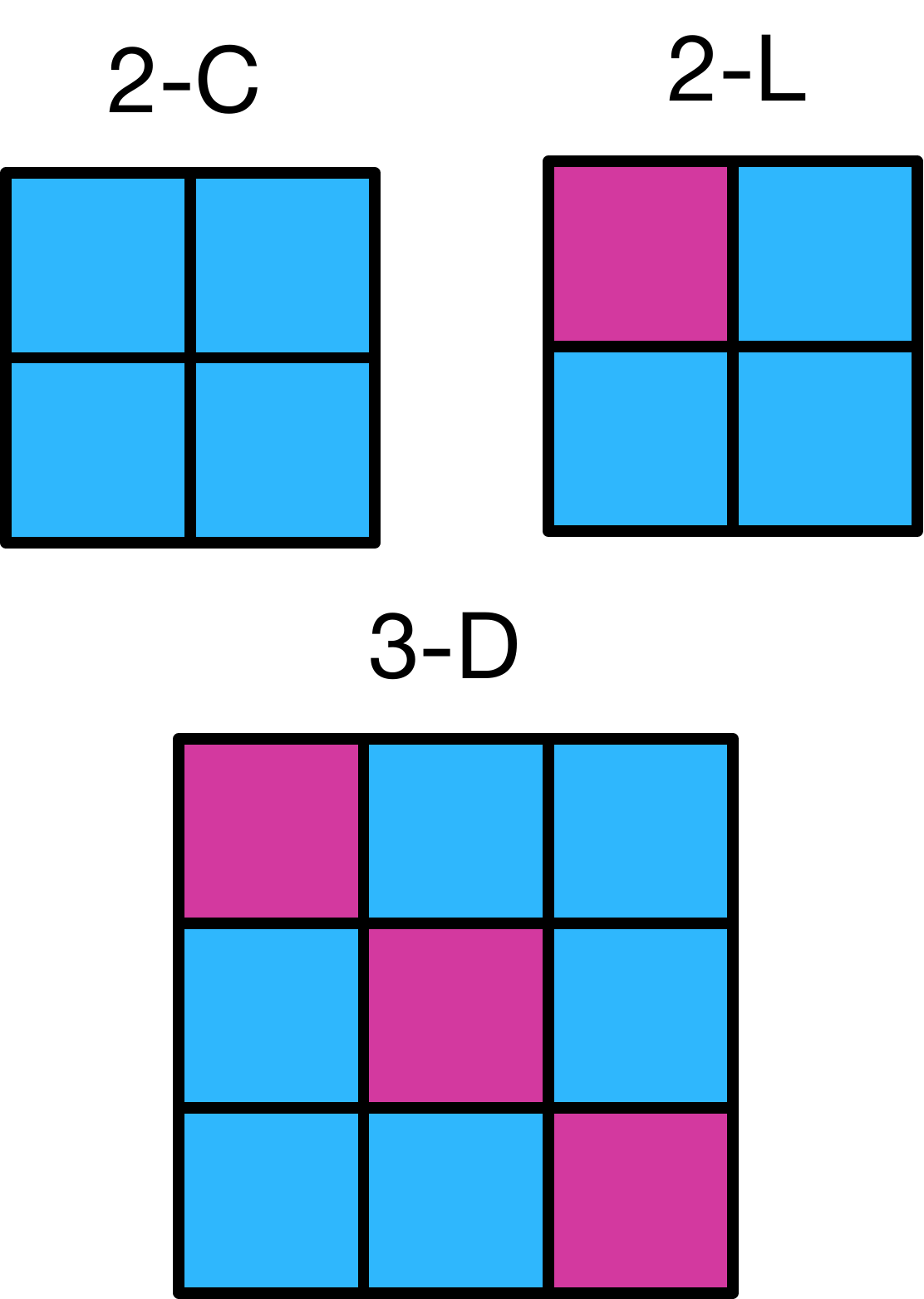}~~
    \includegraphics[width=0.25\linewidth]{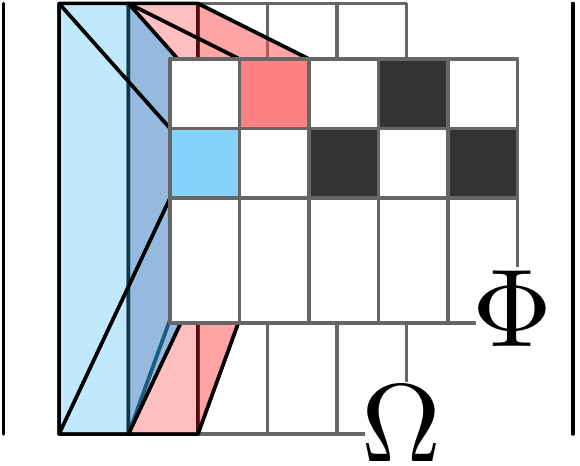}
    \includegraphics[width=0.52\linewidth]{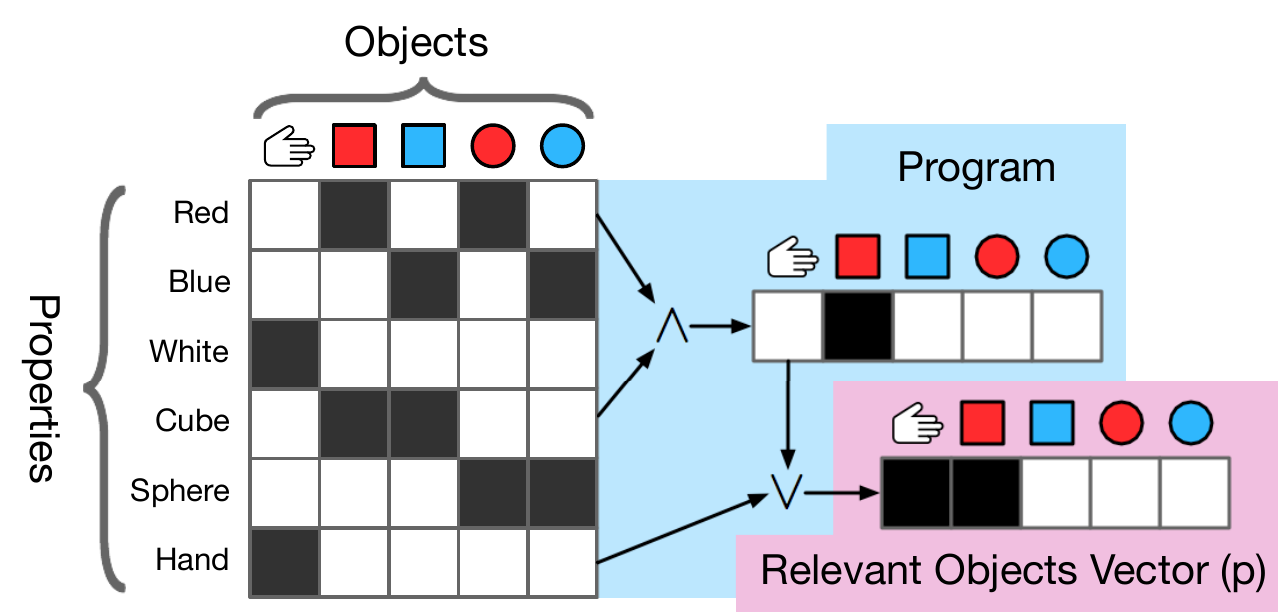}
    \caption{\textbf{Left:} Different designs for 2x2 and 3x3 environments. \textbf{Center:} Relationship between $\vO$ and $\vPhi$.  The features in $\vO$ have detectors applied to them to produce the disentangled representation $\vPhi$.  \textbf{Right:} Example of how the detectors interact with the program.  The property detector outputs $\vPhi$ are shown as binary to ease interpretation, but in practice they are learned continuous values.} % The example program here is to reach to the red cube, which involves both the red cube block and the hand as relevant objects: \texttt{OR(HAND, AND(RED, CUBE))}.}
    \label{fig:program-example}
\end{figure}

% The role of the program in our agent is to allow the network to identify the set of task relevant objects in the environment.  For a reaching task there are two relevant objects: the hand of the robot and the target block the arm is supposed to reach for.  Objects in the environment are identified by a collection of properties that are referenced by the program.

We begin by explaining how the program operates when given a mapping between objects and properties, denoted by $\vPhi$ and explained below.
In later sections we expand on this to explain 
how the assignment of properties to objects can be learned,
and 
how the assignment process can be made differentiable 
to enable end to end training
%in an end to end way 
from a behavioural objective; however, for clarity of presentation it is easiest to begin under the simplifying assumption that the assignments of properties to objects are crisp (i.e.\ 0-1) and known.

The input to the program is a matrix $\vPhi$ whose columns are objects and rows are properties.  The elements of this matrix are in $\{0, 1\}$ (this will be relaxed later) where $\vPhi_{ij} = 1$ indicates that the object $j$ has property $i$.

Figure~\ref{fig:program-example} shows how the set of relevant objects are identified for the example reaching program in Equation~\ref{eq:program-example}.  In this example the environment contains four blocks and the robot hand, for a total of five objects.  Each object has two properties, a color and a shape, which are together enough to uniquely identify it.  The set of \emph{relevant objects} for this program can be expressed as
\begin{align}
    \texttt{OR(HAND, AND(RED, CUBE))}
    \label{eq:relevant-objects-example}
\end{align}
and an indicator function for this set can be obtained by combining the rows of $\vPhi$, as explained below.

Each row of $\vPhi$ corresponds to a particular property that can be referenced in a program, and the values in the rows serve as indicator functions over subsets of objects that have the corresponding property.  These can be used to select new groups of objects by standard Boolean operations, which can be implemented by applying elementwise operations to the rows of $\vPhi$.

Figure~\ref{fig:program-example} shows how the set in Equation~\ref{eq:relevant-objects-example} is obtained.  The functions \texttt{AND} and \texttt{OR} in the specification (shown as $\wedge$ and $\vee$ in the figure) correspond to the set operations of intersection and union, respectively.  The result is a vector whose elements form an indicator function over objects.  The set corresponding to the indicator function contains both the robot hand and the red cube and excludes the remaining objects.

We call the result of this operation the \emph{relevant objects vector} and denote it with $p$ (for ``presence'' in the set of relevant objects).  This vector will play a role in the down stream reasoning process of our agents.

The order of rows and columns of $\vPhi$ is arbitrary.  We take advantage of this to assign indexes to named properties in an arbitrary (but fixed) order.  This is the same type of assignment that is done for language models when words in the model vocabulary are assigned to indexes in an embedding matrix, and imposes no loss of generality beyond restricting our programs to a fixed ``vocabulary'' of properties.

Note that none of the operations described in this section depend on the number of objects.  We will take advantage of this flexibility in the experiments to show that the behaviors we learn can generalize to different numbers of objects in the environment.

\subsection{Differentiating through program execution}

The program execution described in the previous section makes use of set operations on indicator functions, which are uniquely defined when the sets are crisp; however, this uniqueness is lost if the sets are soft.  We would like to apply programs to soft sets so that the assignment of properties to objects can be learned by backprop.  This requires not only that our set operations apply to soft sets, but also that they be differentiable.

There are many ways to meet both of these requirements; for our purposes it is convenient to choose the following assignment:
\begin{align}
    \operatorname{not}(x) = 1-x \qquad
    \operatorname{and}(x, y) = x y \qquad
    \operatorname{or}(x, y) = x+y-xy
    \label{eq:logic}
\end{align}
It can be verified that these operations are self-consistent (e.g.\ $x = \NOT(\NOT(x))$), and reduce to standard Boolean operations when $x, y \in \{0, 1\}$.  This particular assignment is convenient because each operation always gives non-zero derivatives to all arguments.

\subsection{Learned grounding}

The discussion in Section~\ref{sec:executing-programs} assumed that the assignment of properties to objects is given; however, it is much more interesting if our agents can learn to create the matrix $\vPhi$ rather than having it provided by the environment.  In our architecture learning to populate the elements of $\vPhi$ is the role of \emph{detectors}.

A detector operates on a matrix of features $\vO$.  Similar to $\vPhi$, the columns of $\vO$ correspond to objects but the rows of $\vO$ are opaque vectors, populated by whatever information the environment provides about objects.

We create one detector for each property in our vocabulary.  Each detector is a small neural network that maps columns $\omega_j$ of $\vO$ to a value in $[0, 1]$.  Detectors are applied independently to each column of the matrix $\vO$, and each detector populates a single row of $\vPhi$.  Groups of detectors corresponding to sets of mutually exclusive properties (e.g.\ blocks can only have one color in our experiments) have their outputs coupled by a softmax.  For example, when learning the matrix in Figure~\ref{fig:program-example} each column is the output of two softmaxes, one over colors and one over shapes.

These columns of $\vO$ are filled with whatever features the environment provides, position, orientation, etc.  These features must have enough information to identify the properties in our vocabulary, but this information is permitted to be entangled with other features in $\vO$, while it must be disentangled in $\vPhi$.  This relationship is diagrammed in Figure~\ref{fig:program-example}.

It would be simple to pre-train the detectors for each property and provide those to the agent.  However, one of the contributions of this paper is to show that with our network architecture we do not need to do this.  Our agents learn to identify meaningful properties of objects and to reason about sets of objects formed by combinations of these properties in a \emph{completely end to end way}.

\subsection{Relational reasoning}

Up to this point we have described completely separate processing for each object.  The agent receives a matrix $\vO$ whose rows are features and columns are objects.  We apply a battery of detectors to each column $\vO$ to create the matrix $\vPhi$ where rows are properties and columns are again objects.  The program then applies elementwise operations to the rows of $\vPhi$ to create the relevant objects vector $p$.

In order to allow reasoning over relationships between objects we introduce a message passing scheme to exchange information between the objects selected by the relevant objects vector.  Our message passing operation closely resembles an Interaction Network~\cite{Battaglia2016a} with some additional features.

Using $\omega_i$ and $\omega_j$ to represent columns of $\vO$, we can write a single round of message passing as
\begin{align}
    \omega'_i = f(\omega_i) + \sum_j \alpha_{ij} r(\omega_i, \omega_j)
    \label{eq:message-passing}
\end{align}
where $\omega'_i$ is the resulting transformed features of object $i$.  This operation is applied to each column of $\vO$, and the resulting vectors are aggregated into the columns of a new matrix $\vO'$.  The function $f(\omega_i)$ produces a local transformation of the features of a single object, and $r(\omega_i, \omega_j)$ provides a message from object $j \to i$.
We implement the functions $f$ and $r$ with small MLPs, and structurally the message passing operation is similar to that of Interaction Networks~\cite{Battaglia2016a} and the Neural Physics Engine~\cite{chang2017}.

Messages between objects are mediated by edge weights $\alpha_{ij}$ which are determined using a modified version of the \emph{neighborhood attention} operation of Duan~et~al.~\cite{Duan2017},
\vspace{-0.1em}
\begin{align*}
    c_i &= \operatorname{Linear}(\omega_i) & q_i &= \operatorname{Linear}(\omega_i) 
     & \tilde{\alpha}_{ij} &= w^\T\tanh(q_i + c_j) & \alpha_{ij} &= \frac{p_j \exp{\tilde{\alpha}_{ij}}}{\sum_k p_k \exp{\tilde{\alpha}_{ik}}}
\end{align*}
Recall that $p$ is the relevant objects vector (which is a result of the program, see Figure~\ref{fig:program-example}), and that its elements lie in the interval $[0, 1]$.  To understand why they are included in this way, consider what happens if $p_j = 0$, which means that object $j$ is not a relevant object for the current task.  In this case the resulting $\alpha_{ij} = 0$ also, and the effect is that the message from $j \to i$ in Equation~\ref{eq:message-passing} does not contribute to $\omega_i'$.  In other words, task-irrelevant objects do not pass messages to task-relevant objects during relational reasoning.

If we were to consider different types of relations in our programs (beyond \texttt{NEAR}), we would do so by parameterizing the message passing operation based on the relation to be achieved.  Since we consider only reaching tasks, this extra complexity is not needed in the present work.

After message passing an output can be produced by computing $a = \operatorname{MLP}(\tanh(\langle\vO', p\rangle))$.  For full details see Appendix~\ref{sec:training-details}.

%% file: 060-experiments.tex
\section{Experiments}

\begin{figure}[tb]
    \centering
    \includegraphics[width=0.238\linewidth]{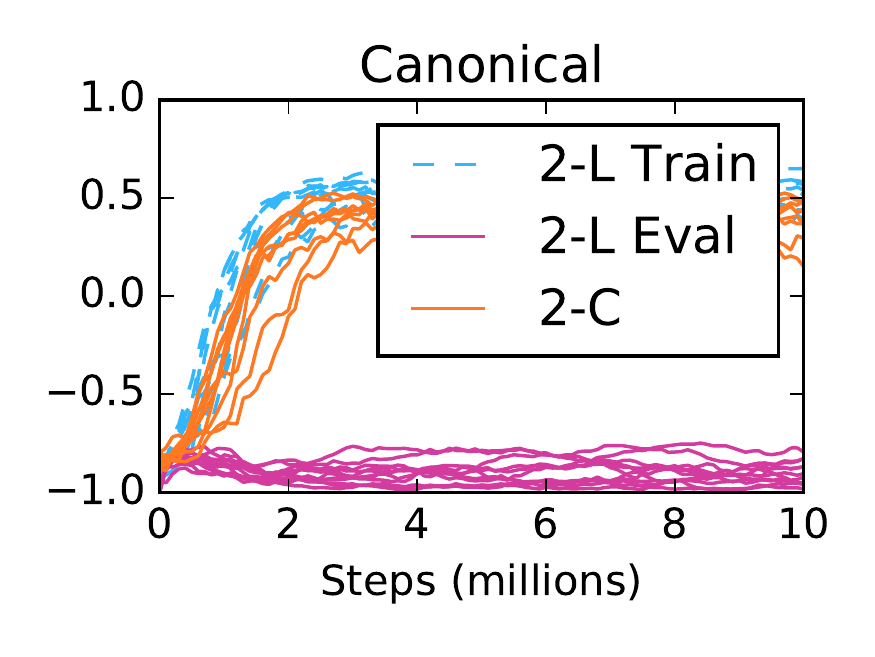}~~
    \includegraphics[width=0.238\linewidth]{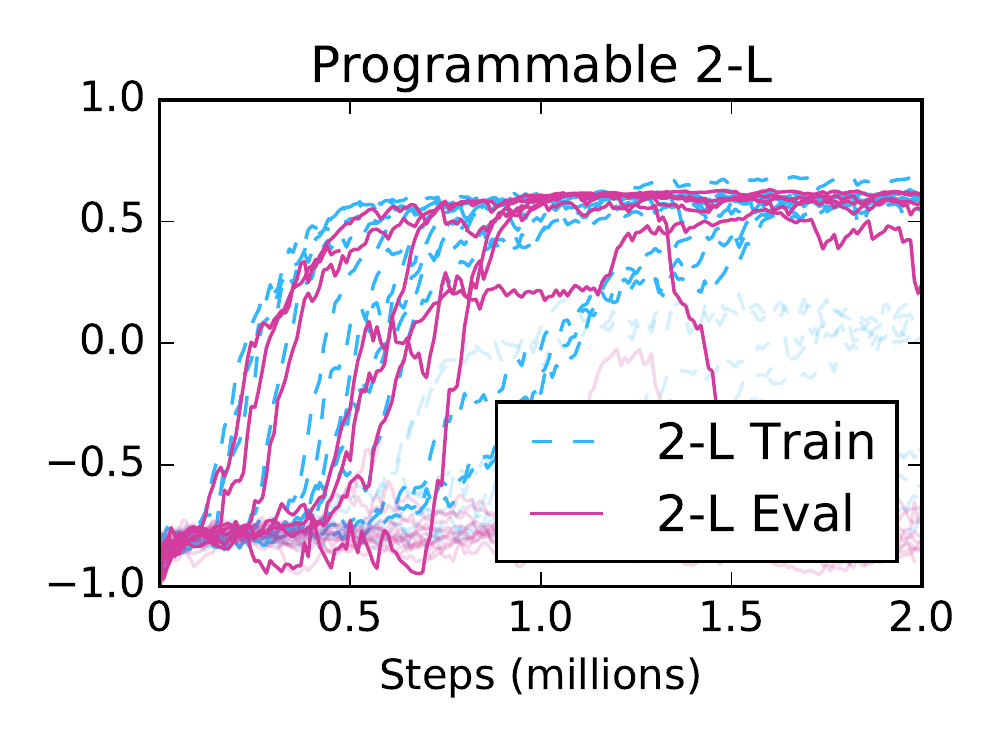}
    \includegraphics[width=0.238\linewidth]{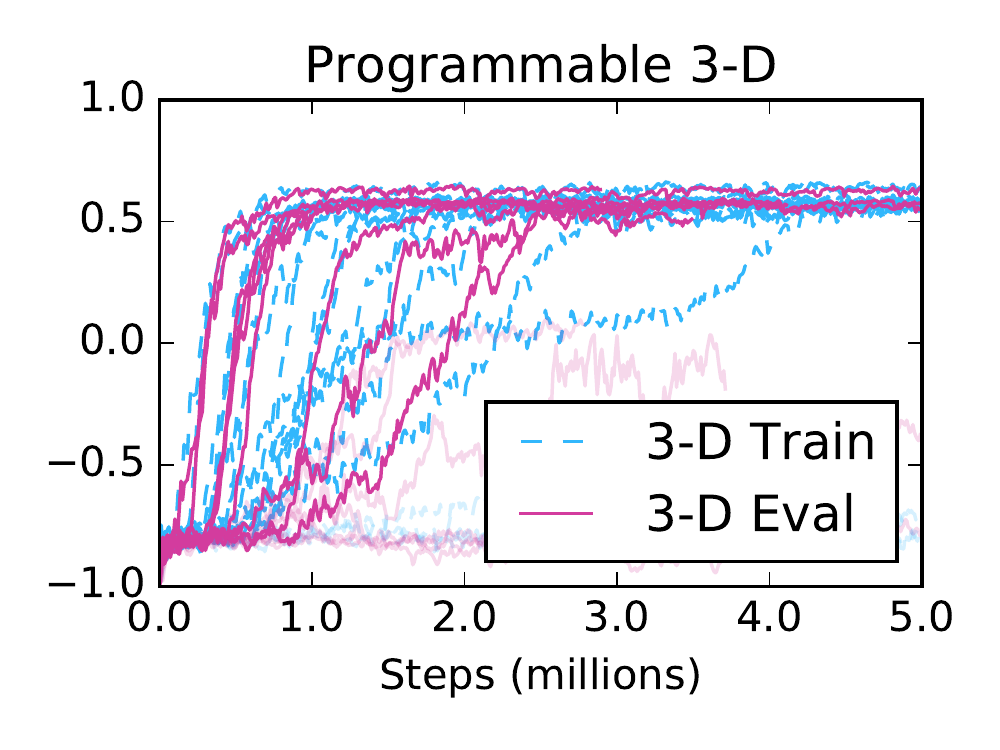}~~
    \includegraphics[width=0.238\linewidth]{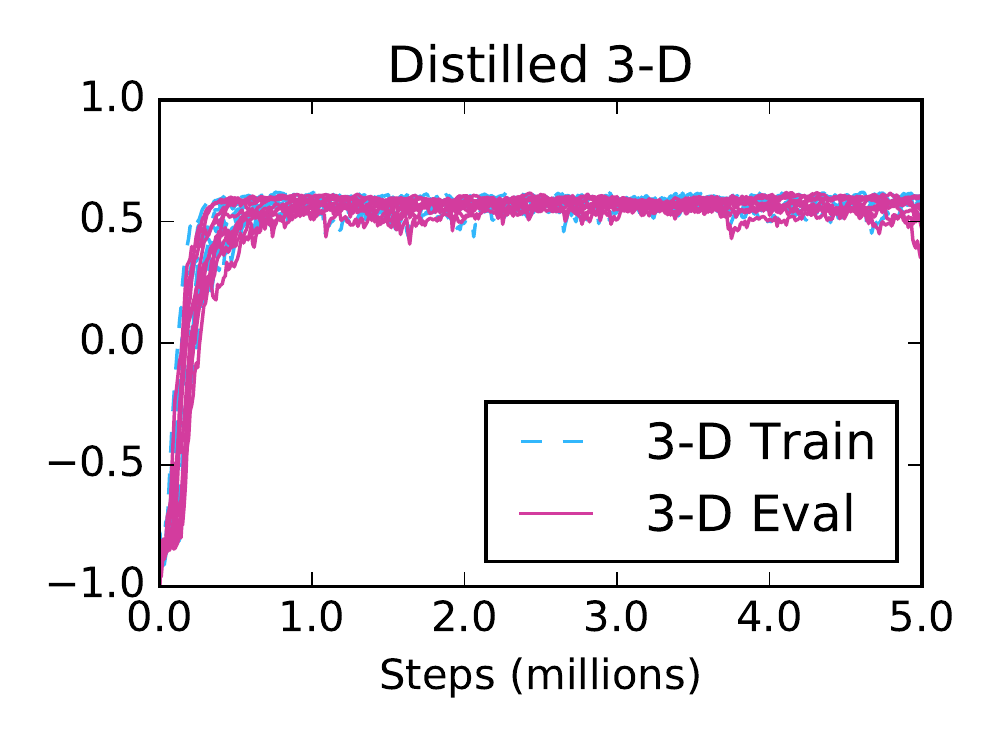}
    \caption{\textbf{Left:} Performance of the canonical and programmable architectures on the 2-C and 2-L environment variants. \textbf{Center:} Training and generalization curves for programmable agents with learned detectors trained on the 2-L and 3-D reaching variants.  Learning is somewhat unstable in these cases, but successful agents learn to generalize perfectly. \textbf{Right:} Training and generalization curves for distilled agents trained on the 3-D variant (results for 2-L are similar). Exploiting ground truth property information at training time substantially stabilizes learning.}
    \label{fig:learned-detector-results}
\end{figure}

In this section we look at various ways that agents implemented as Programmable Networks can generalize.  Throughout this section we refer to different variants of the programmable reaching environment using the names indicated in the left panel of Figure~\ref{fig:program-example}.  All of our agents are trained using Deterministic Policy Gradient~\cite{silver2014deterministic, lillicrap2015continuous}, where both the actor and the critic are Programmable Networks.  See Appendix~\ref{sec:training-details} for full details of our training setup.

\subsection{Standard methods fail to generalize}
\label{sec:standard-failure}

We start by showing that the programmable reaching environment poses a non-trivial challenge.  This is perhaps not immediately obvious, because the individual tasks (i.e.\ reaching for a single block) are very simple.  The 2-C design, where all possible target blocks are seen during training, is readily solved by standard deep architectures.  Where standard architectures fail (and do so catastrophically) is in generalization to unseen combinations of properties.

For the standard architecture we compare against an architecture similar to the one of Lillicrap~et~al.~\cite{lillicrap2015continuous}, modified for feature based observations.  We call this the ``canonical'' architecture as it has been shown to achieve good performance on many continuous control tasks.

% \begin{figure}
%     \centering
%     \includegraphics[width=0.24\linewidth]{figures/canonical-2x2}
%     \includegraphics[width=0.24\linewidth]{figures/programmable-2L-oracle}
%     \caption{Performance of the canonical and programmable architectures on the 2-C and 2-L environment variants.}
%     \label{fig:canonical-failure}
% \end{figure}

Performance of the canonical architecture on the 2-C and 2-L designs is shown in the left plot of Figure~\ref{fig:learned-detector-results}.  Reward per step of \textasciitilde 0.6 is effectively perfect performance in these environments.

What we see from these curves is that the canonical architecture achieves good performance on all of the tasks it was trained on, but in the 2-L case it completely fails to generalize its behavior to reaching for the held out target.  Good performance on 2-C verifies that this is not merely a capacity problem; when the network is trained on all four tasks it is able to achieve all four.

\subsection{Zero-shot tasks}

The center plots in Figure~\ref{fig:learned-detector-results} show performance curves for the Programmable Agent on the training and evaluation conditions of the 2-L and 3-D designs.  
Learning is not always stable, but is nonetheless successful in many cases.  We emphasize that the difficulty here is stability rather than a performance; agents that succeed in training perform very well in generalization.

Stability of training can be improved by taking advantage of ground truth property information in the critic at training time.  In this setting the actor still learns detectors, but the critic does not.  We call this the ``distilled'' model, and learning curves for distilled agents on the 3-D design are shown in the right of Figure~\ref{fig:learned-detector-results}.

% We can improve the stability of learning by taking advantage of oracle information at training time.  Recall that our agents have both an actor and a critic network, but only the actor is needed to act in the environment.
% Using ground truth property information in the critic (but not the actor) makes learning very stable. This is shown in Figure~\ref{fig:learned-detector-results} (right).
% In this case the behavior is still generated by the actor, which must rely on its learned detectors to identify the properties of objects.  It is only the critic (which is used to train the actor) that takes advantage of oracle information.  We call this combination of learned detectors in the actor and oracle information in the critic the ``distilled'' agents.

% For the remaining experiments in this paper we select one of the successful agents with learned detectors (i.e.\ not distilled) and further probe its generalization capabilities in a variety of ways.

\subsection{Strong generalization}
\label{sec:generalization}

\begin{figure}[tb]
    \centering
    \includegraphics[width=\linewidth]{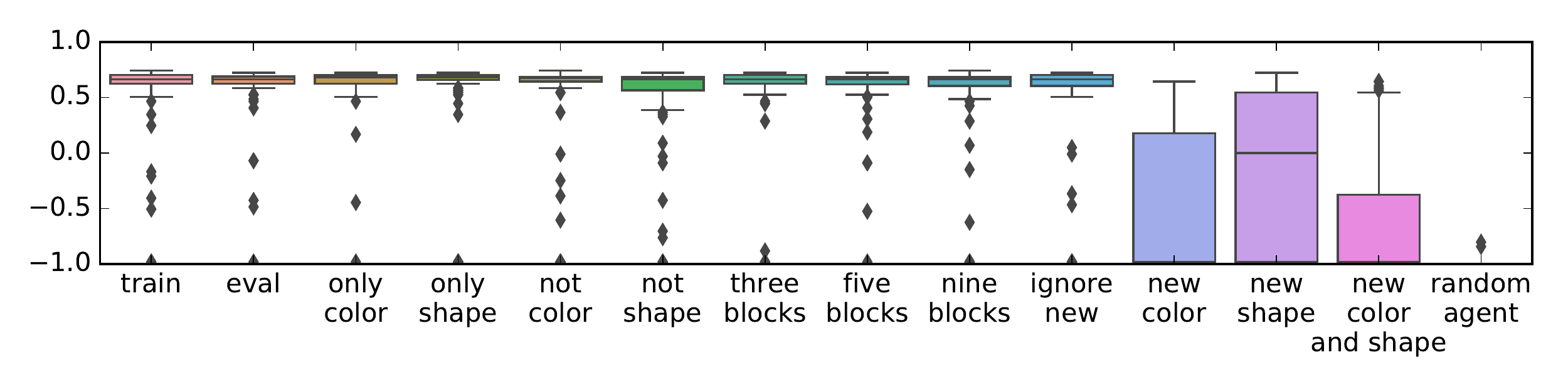}
    \vspace{-25px}
    \caption{Generalization experiments.  Each column shows the performance distribution of a single agent on zero-shot tasks under different conditions.  See text for a full description.}
    \label{fig:generalization}
    \vspace{-5px}
\end{figure}

\begin{figure}[tb]
    \centering
    \includegraphics[width=0.45\linewidth]{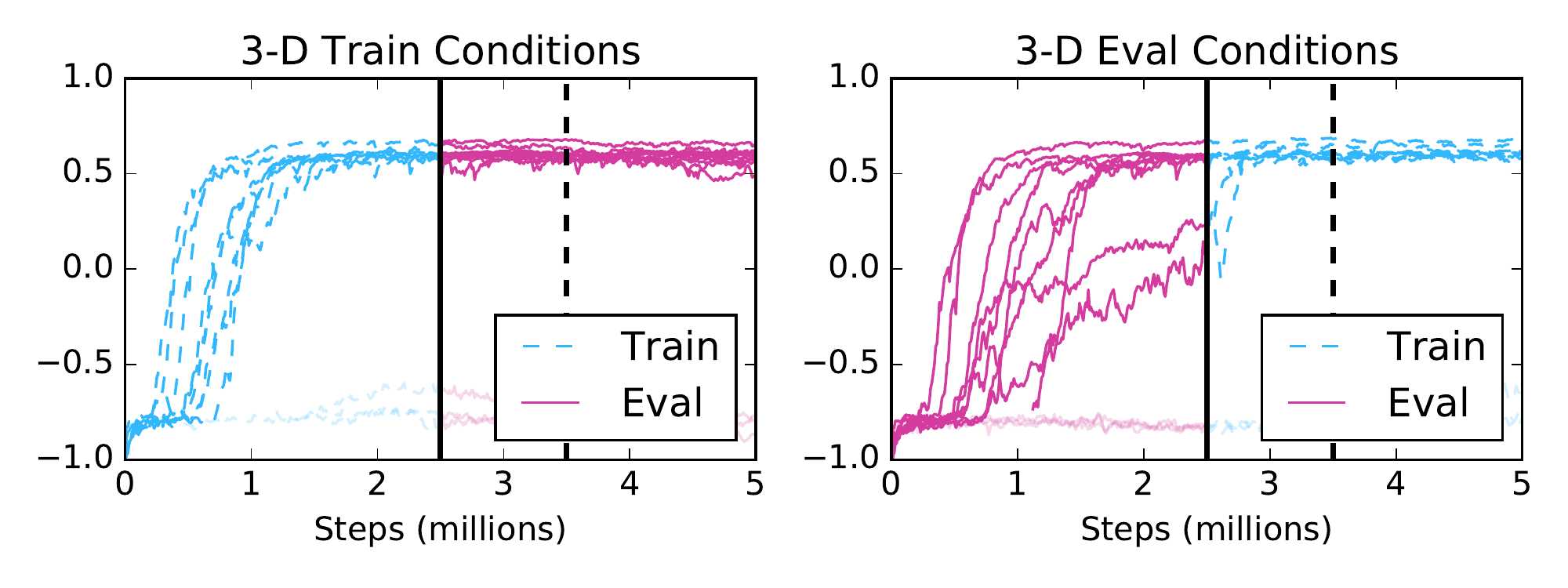}~~~
    \includegraphics[width=0.525\linewidth]{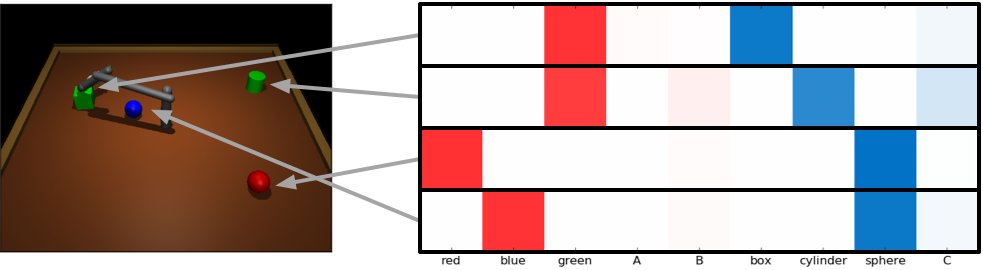}
    \vspace{-15px}
    \caption{\textbf{Left:} performance on the off-diagonal and diagonal tasks for the 3x3 environment. Training begins on the off-diagonal tasks and switches to the diagonal tasks at 2.5m steps (marked by the solid black line). The dashed black line shows the point where all transitions from off-diagonal task have been pushed out of the replay buffer.  \textbf{Right:} The learned property matrix $\vPhi$ (actually its transpose) produced from the adjacent environment state.  See Appendix~\ref{sec:detectors} for additional visualizations.}
    \label{fig:detectors}
    \vspace{-10px}
\end{figure}

Because our architecture partitions the objects into sets that have different properties we can achieve very strong generalization. Figure~\ref{fig:generalization} shows zero-shot performance of one of our agents with fully learned detectors (i.e.\ no distillation) trained on the 3-D design.  Each column summarizes the reward per step achieved by the agent over 100 testing episodes under different generalization conditions. 

Figure~\ref{fig:generalization} also shows two baseline conditions.  The \texttt{train} condition corresponds to the agent being tested on new episodes of the tasks it was trained on, and the \texttt{random} condition corresponds to the performance of the same agent with random weights.

We only show results for the Programmable Agent because we were not successful in training agents with the canonical architecture on any of 3x3 design.  Even when all tasks are seen during training (the most informative possible 3x3 setting) the canonical architectures were no better than random.

The zero-shot conditions in Figure~\ref{fig:generalization} are as follows:\footnote{Videos for each condition can be found at: \hideurl{https://goo.gl/UkvWYE}}\vspace{-0.15em}
\begin{itemize}
    \item \texttt{eval} shows performance on the zero-shot tasks of the 3-D design, as shown in Figure~\ref{fig:program-example}.
    \item \texttt{only color/shape} shows performance when only the color or shape is specified.
    \item \texttt{not color/shape} shows performance when the color or shape to \emph{not} reach for is specified.
    \item \texttt{\# blocks} show performance of zero-shot tasks with different numbers of blocks on the table.  All training episodes had exactly four blocks.
    \item \texttt{ignore new blocks} shows the performance of zero-shot tasks in the presence of new block shapes and colors.
    \item \texttt{reach new shape} shows the performance when reaching for blocks with a novel shape and known color.
    \item \texttt{reach new color} shows the performance when reaching for blocks with a novel color and known shape.
    \item \texttt{reach new shape and color} shows the performance when reaching for a block when both the color and shape are novel.
\end{itemize}
We introduce new blocks by adding an additional shape and color to the set of possible blocks that can appear on the table (for a total of 16 possible combinations).  In the \texttt{ignore new blocks} condition blocks with the new properties can appear on the table as distractors, but are never the target of a reaching program.  In the \texttt{only shape} and \texttt{only color} conditions we modify the environment generation process to ensure that the target is uniquely identified by the specified color or shape.

Apart from the \texttt{train} and \texttt{random} conditions, every episode of every task in Figure~\ref{fig:generalization} is zero-shot.  Every program specifies the target in a way the agent did not see at training time.  When showing generalization to different numbers of blocks (in the \texttt{\# blocks} conditions) and the ability to ignore novel distractor blocks (in \texttt{ignore new blocks}) we are showing the agent reaching for a zero-shot target under these conditions.

% Targeting novel colors and shapes is done via the principle of exclusion.  For example, let us say we have five color detectors labelled \texttt{[RED, GREEN, BLUE, A, B]} where \texttt{A} and \texttt{B} are never seen at training time.  At test time we can represent the set of objects of novel color by computing  \texttt{OR(AND(NOT(RED), NOT(GREEN), NOT(BLUE)), A, B)}.  Novel shapes can be specified in a similar way.

% This method of identifying novel levels of properties highlights a more subtle type of generalization that is taking place.  We specified three logical operations in Equation~\ref{eq:logic}, $\operatorname{and}$, $\operatorname{or}$ and $\operatorname{not}$; but we train the agent on programs that only use $\operatorname{and}$ and $\operatorname{or}$.  We can use the $\operatorname{not}$ we have defined at test time because it is \emph{compatible} with the other two operations, in the sense that $\operatorname{or}(x, y) = \operatorname{not}(\operatorname{and}(\operatorname{not}(x), \operatorname{not}(y)))$.  This compatibility is enough to ensure that the $\operatorname{not}$ operation imposes the intended semantics on the agent representations, in spite of the fact that no program with a $\operatorname{not}$ operation ever appeared during training.

The right panel of Figure~\ref{fig:detectors} shows the learned detector outputs in a successful agent.  This figure shows that the agent can correctly identify properties of objects referenced by the programs it executes.

\subsection{No catastrophic forgetting}

Our agents are robust against catastrophic forgetting.  To show this we trained a collection of agents on the 3x3 environment for 5m steps.  For the first 2.5m steps we follow the train condition of the 3-D design, and for the remaining 2.5m steps we switch to training on the test conditions.  The plots in Figure~\ref{fig:detectors} show performance curves for these agents on both the training and test conditions throughout the full 5m steps of training.  Performance on the train condition does not degrade even millions of steps after the switching to training on new tasks.

% \begin{figure}
%     \centering
%     \includegraphics[width=0.9\linewidth]{figures/no-catastrophic-forgetting}
%     \caption{These plots show performance on the off-diagonal and diagonal tasks for the 3x3 environment. Training begins on the off-diagonal tasks and switches to the diagonal tasks at 2.5m steps (marked by the solid black line).  After switching the off-diagonal tasks are not seen again by the training process.  The dashed black line shows the point all transitions from off-diagonal task have been pushed out of the replay buffer.}
%     \label{fig:no-catastrophic-forgetting}
% \end{figure}

%% file: 070-conclusion.tex
\section{Conclusion}

Our Programmable Network architecture enables us to build agents that execute declarative programs expressed in formal language.  Our agents learn to ground the terms of the programs in their environment and can leverage these grounded terms to generalize beyond the tasks they were trained on.  Our agents achieve nearly perfect generalization on a variety of zero-shot tasks where standard deep RL architectures completely fail.

All of this is achieved with only small concessions from the Deep Learning zeitgeist.  We assume that the boundaries between objects in the agent observations are known, and the object properties our agents can reason over must come from a predefined vocabulary, which is similar to the restriction one has in word based language modelling.  We also use a fixed mapping from programs to architecture.

Future work will focus on improving the robustness of training, and on scaling the method to deal with more types of properties (e.g.\ mass, size, texture, etc) and relations.  We are also interested in extending this method to work from vision.
%\note{expand this paragraph to talk about relations, Last sentence could be: We are also interested in detecting objects and predicting their features directly from pixels and other modalities.}

%% file: 080-training-details.tex
\section{Model and training details}
\label{sec:training-details}

\begin{figure}[H]
    \centering
    \includegraphics[width=0.9\linewidth]{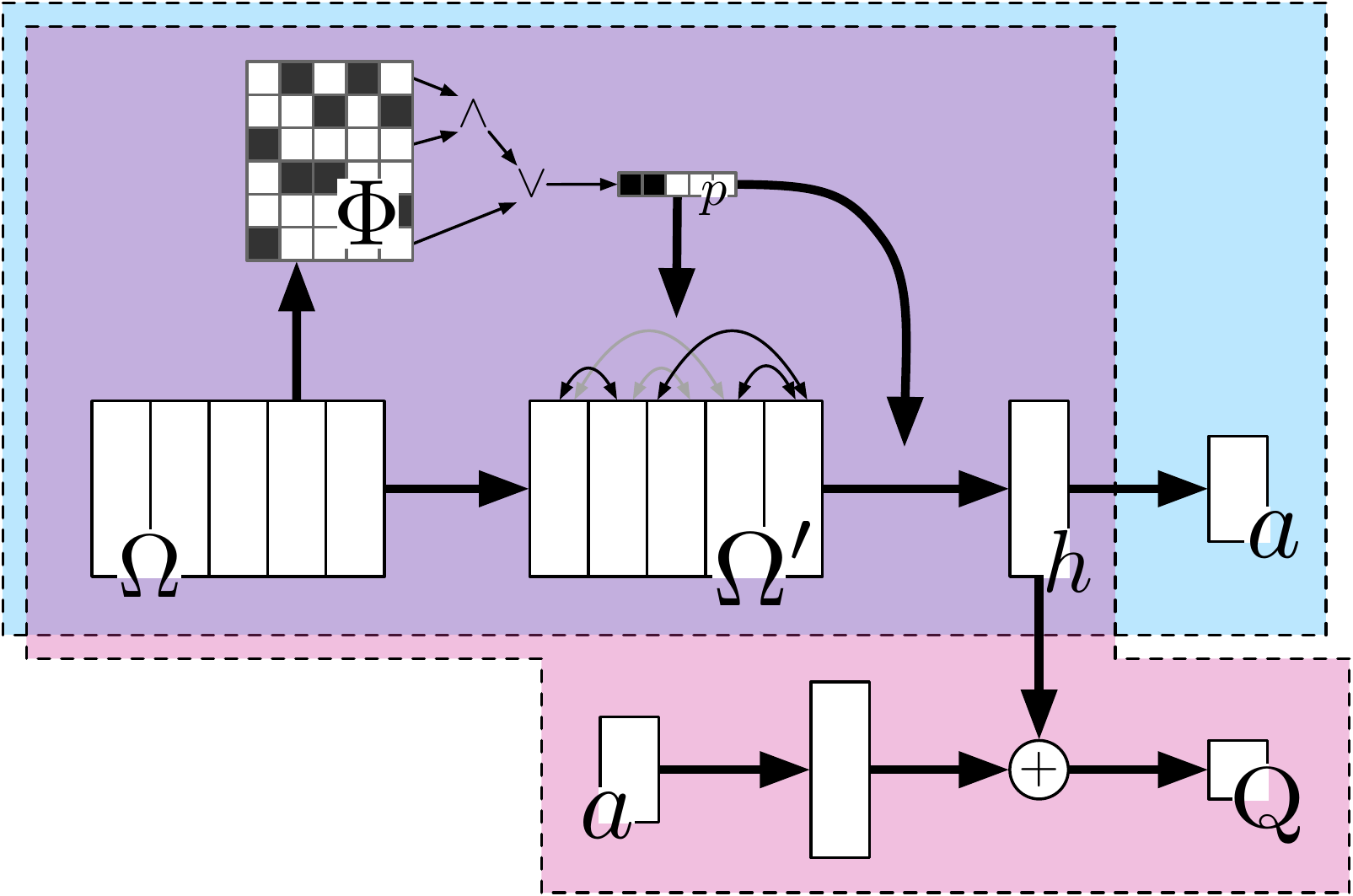}
    % \caption{The full programmable architecture for both the actor and the critic networks.  The blue box encompasses the actor network, and the purple box encompasses the critic.  The actor and critic share the network structure in the overlapping section, but the weights for each network are separate.}
    \label{fig:programmable-architecture}
\end{figure}

Our agents are trained using DPG~\cite{silver2014deterministic, lillicrap2015continuous}, where both the actor and critic are programmable networks.  The full architecture of the actor and critic are shown in Figure~\ref{fig:programmable-architecture}.  The actor and critic share the same programmable structure (including the vocabulary of properties), but they do not share weights.

The relationship between $\vO$, $\vO'$, $\vPhi$ is explained in the main text, as is the derivation of the relevant objects vector $p$.  In both the actor and critic the vector $h$ is produced by taking a weighted sum over the columns of $\vO'$.  Using $\omega_i'$ to denote these columns, we can write $h$ as
\begin{align*}
    h = \sum_i p_i \omega_i'
\end{align*}
The motivation for weighting the columns by $p$ here is the same as for incorporating $p$ into the message passing weights in Equation~\ref{eq:message-passing}; we want $h$ to include only information about relevant objects, and the role of $p$ is precisely to identify these objects.  Reducing over the columns of $\vO'$ fixes the size of $h$ to be independent of the number of objects.

From this point the architectures of the actor and critic diverge.  Recall that there are two networks here that do not share weights, so there are in fact two different $h$ vectors to consider.  We distinguish between the activations at $h$ in the actor and critic by using $h_a$ to denote $h$ produced in the actor and $h_c$ to denote $h$ produced in the critic.

The actor produces an action from $h_a$ using a single linear layer, followed by a $\tanh$ to bound the range of the actions
\begin{align*}
    a = \tanh(\operatorname{Linear}(\tanh(h_a))) \enspace.
\end{align*}

The computation in the critic is slightly more complex.  Although $h_c$ contains information about the observation, it does not contain any information about the action, which the critic requires.  We combine the action with $h_c$ by passing it through a single linear layer which is then added to $h_c$
\begin{align*}
    Q(\vO, a) = \operatorname{Linear}(\tanh(h_c + \operatorname{Linear}(a))) \enspace.
\end{align*}
No final activation function is applied to the critic in order to allow its outputs to take unbounded values.

\subsection{Building the observation matrix}

The observations consumed by our agent are collected into the columns of $\vO$.  The matrix $\vO$ has one column for each object in the environment, where objects include all of the blocks on the table and also the hand of the robot arm.

As discussed in Section~\ref{sec:environment}, each object is described by its 3d position and 4d orientation, represented in the coordinate frame of the hand.  Each block also has a shape and a color which are represented to the agent using 1-hot vectors.  There are 4 possible colors and 5 possible shapes for a total of 9 property features (and 16 total features) for each object.

We also provide the joint positions of the arm as observations (encoded as $\sin$ and $\cos$ of the joint angle).  The arm has 6 joints, for a total of 18 features to represent the arm.

The full observation matrix $\vO$ is built by appending the arm positions to each (object) column, which effectively represents each object in a ``body pose relative'' way.  This means that in an environment with four blocks we have an observation of shape $34 \times 5$ (16 + 18 features per object for each of the 4 + 1 objects). 

In addition to the above, we also provide the agent with the index of the hand in $\vO$, so it does not need to learn to detect its own body.

\subsection{Training parameters}

Each reaching episode lasts 10 seconds and controls are issued at 0.1 second intervals, for a total of 101 observations per episode (including the initial observation at $t=0$).  Our agents are trained for either 2m or 5m frames (in the 2x2 or 3x3 environments, respectively).  We collect experience into a replay buffer with a capacity of 1m frames, which we use for training.  After each action we sample a batch of 64 transitions from the replay buffer which is used to update both the actor and the critic.  We use a target network~\cite{mnih2015human} for both the actor and the critic to stabilize training, and we copy the latest parameters to the target networks after each episode (we do not use the soft updates of Lillicrap~et~al.~\cite{lillicrap2015continuous}).

All agents are trained with a discount factor of $\gamma=0.99$, and when computing updates for the actor we clip each element of the gradient from the critic to the range $[-1, 1]$.  During training we use exploration noise that follows a Ornstein–Uhlenbeck process with standard deviation 0.3 and a damping factor of 1.0.

\begin{wrapfigure}{R}{0.55\textwidth}
  \begin{center}
  \quad
    \includegraphics[width=0.45\linewidth]{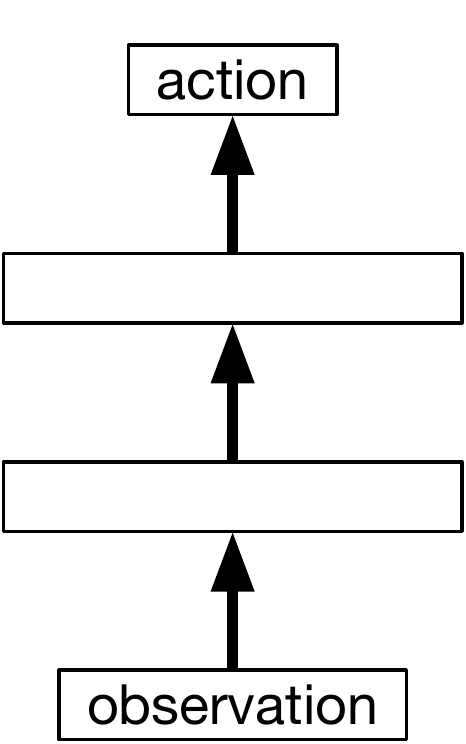}
    \includegraphics[width=0.45\linewidth]{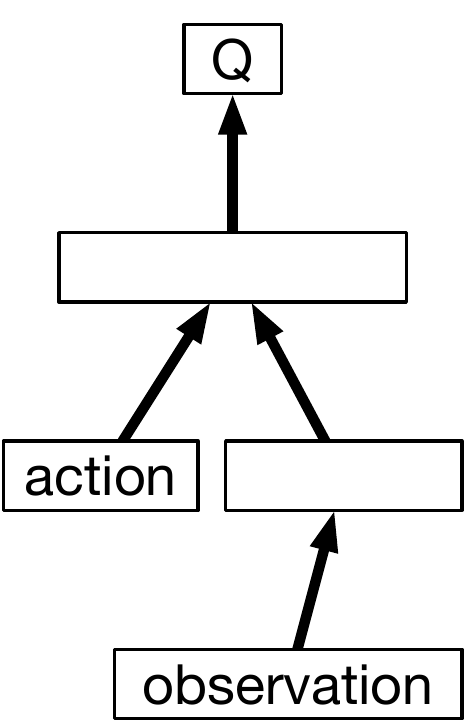}
  \end{center}
  \caption{Structure of the canonical actor and critic networks. Arrows represent linear layers; $\tanh$ activations are used throughout both networks.}
  \label{fig:canonical-networks}
\end{wrapfigure}

The programmable networks use a hidden layer size of 150 units in both the actor and the critic.  The neighborhood attention operation using a context and query size ($c_i$ and $q_i$) with 64 units.  Detectors are implemented as logistic regressions on object features.

The structure of the canonical actor and critic networks is shown in Figure~\ref{fig:canonical-networks}.  In our experiments these networks have 400 units in each hidden layer and use $\tanh$ activations throughout.  Their weights were initialized following the strategy of Lillicrap~et~al.~\cite{lillicrap2015continuous}.  When training the canonical networks we used the same environment parameters as for the programmable networks.  Settings for discounting, target networks and exploration noise were also the same.  The canonical networks accept vectors of observations (rather than the matrices we feed to the Programmable Networks).  We obtain observations for the canonical networks by flattening the $\vO$ matrix into a vector, ensuring that objects with the same properties always appear at the same locations in the flattened vector.

%% file: 090-novel-objects.tex
\section{Referencing objects by exclusion}
\label{sec:exclusion}

% \setlength{\epigraphwidth}{.5\textwidth}
% \epigraph{\textit{``Once you eliminate the impossible, whatever remains,
% no matter how improbable, must be the truth.''}}{\textit{Sherlock Holmes}}

\begin{quote}\textit{Once you eliminate the impossible, whatever remains,
no matter how improbable, must be the truth.} --- \textit{Sherlock Holmes}
\end{quote}

Referencing objects by properties they do not have (e.g.\ ``the cube that is not red'') works by exclusion.  To reach for an object without a property we can simply write a program that expresses this.  The program
\begin{align*}
    \texttt{NEAR(HAND, AND(NOT(RED), CUBE))},
\end{align*}
directs the agent to reach for the cube that is not red.

This method of referencing the absence of properties highlights yet another type of generalization that is taking place.  We specified three logical operations in Equation~\ref{eq:logic}, $\operatorname{and}$, $\operatorname{or}$ and $\operatorname{not}$; however, training programs are all of the form
\begin{align*}
    \texttt{NEAR(HAND, AND(shape, color))},
\end{align*}
which do not make use of the $\NOT$ operation.  Nonetheless, agents are still capable of executing programs that contain negations.

The reason this works is that the operations in Equation~\ref{eq:logic} are \emph{compatible}.  DeMorgan's laws require that negation interact with $\AND$ and $\OR$ in a particular way, and the rules of classical logic require that these laws hold.  The definition of $\NOT$ we have chosen satisfies these relations, and this is enough for the $\NOT$ operation to acquire negation semantics in our networks.

Referencing novel colors and shapes works in a similar way.  For example let us say we have a vocabulary of five colors (the procedure for referencing novel shapes is completely analogous), but only three of them have appeared in the training data.  We can label the colors
\begin{align*}
    \texttt{[RED, GREEN, BLUE, A, B]},
\end{align*}
where we use generic names like a \texttt{A} and \texttt{B} to stand in for color terms that have never been used (and so, in principle, could be anything).  In this case we can express the concept of ``novel color'' in two ways.  The first is an exclusive expression,
\begin{align*}
    \texttt{NOT(OR(RED, BLUE, GREEN))},
\end{align*}
which says ``not any of the colors that have appeared,'' and the second is an inclusive expression,
\begin{align*}
    \texttt{OR(A, B)},
\end{align*}
which says ``any of the colors that have not appeared.''  In practice we have found that combining both methods
\begin{align}
    \texttt{OR(NOT(OR(RED, BLUE, GREEN)), OR(A, B))}
    \label{eq:exclusive-reference}
\end{align}
to give the best performance, since it leverages our model's assumption that every object has exactly one color (i.e.\ the soft membership values for all color sets must sum to 1).

Using the technique of Equation~\ref{eq:exclusive-reference} we can write a program to reach for the block with a new shape and a new color as
\begin{center}
\begin{minipage}{.75\textwidth}
\begin{verbatim}
NEAR(
    HAND,
    AND(
        OR(NOT(OR(RED, BLUE, GREEN)), OR(A, B)),
        OR(NOT(OR(CUBE, SPHERE, CYLINDER)), C)
    )
)
\end{verbatim}
\end{minipage}
\end{center}

The programs referencing novel objects are much more complex than the training programs, but agents perform much better than chance in these cases as well.

%% file: 110-detector-visualizations.tex
\section{Detector output visualizations}
\label{sec:detectors}

The figures show an environment state along with the corresponding $\vPhi$ (transposed, so objects are in rows).  Columns corresponding to different properties have been annotated with their corresponding color and shape terms.  Generic names (\texttt{A, B, C}) indicate un-grounded concepts.

\subsection{Known properties}

\begin{center}
\includegraphics[width=0.92\linewidth]{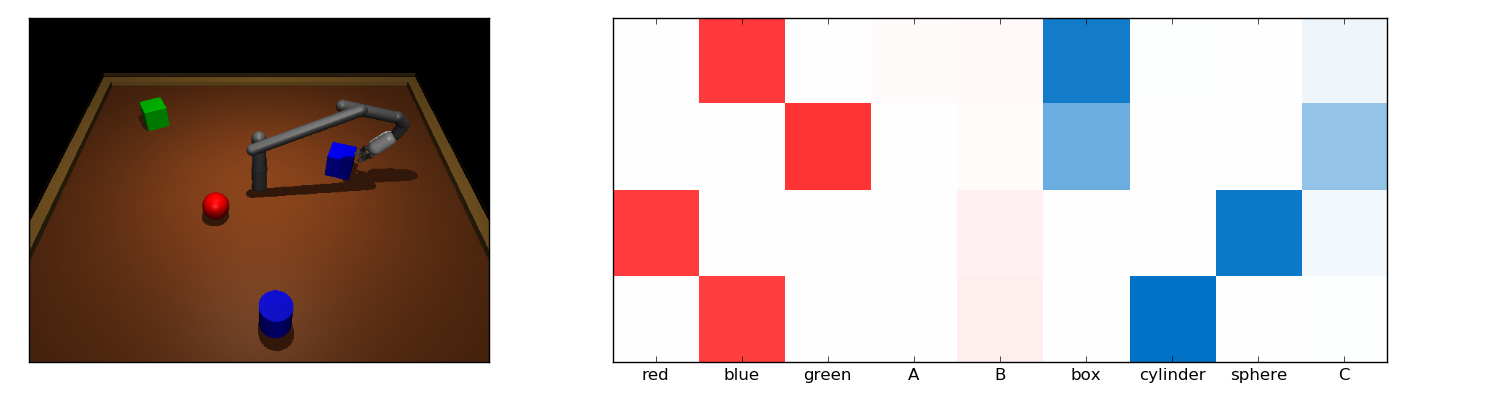}
\includegraphics[width=0.92\linewidth]{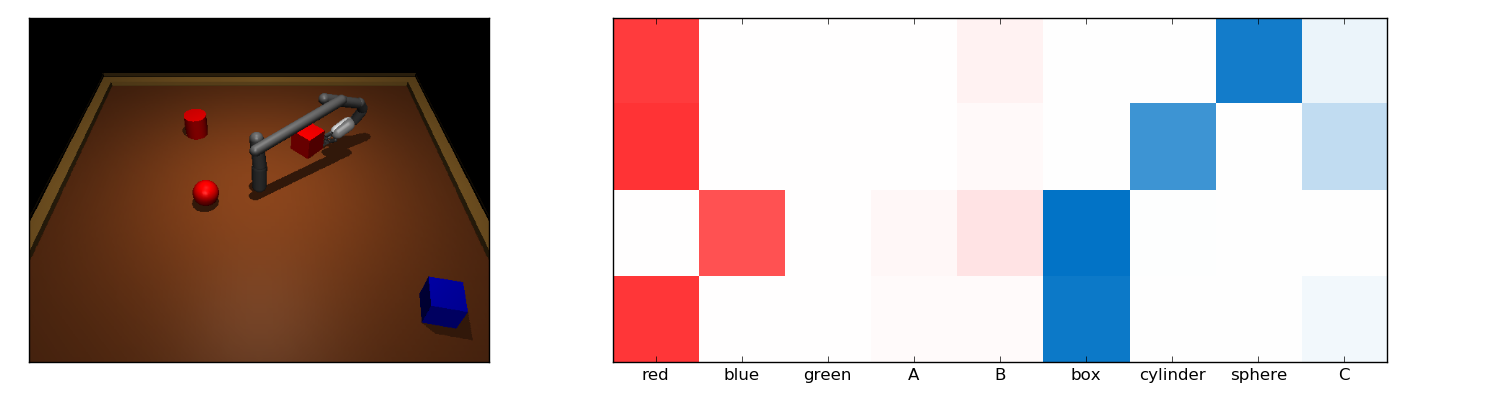}
\includegraphics[width=0.92\linewidth]{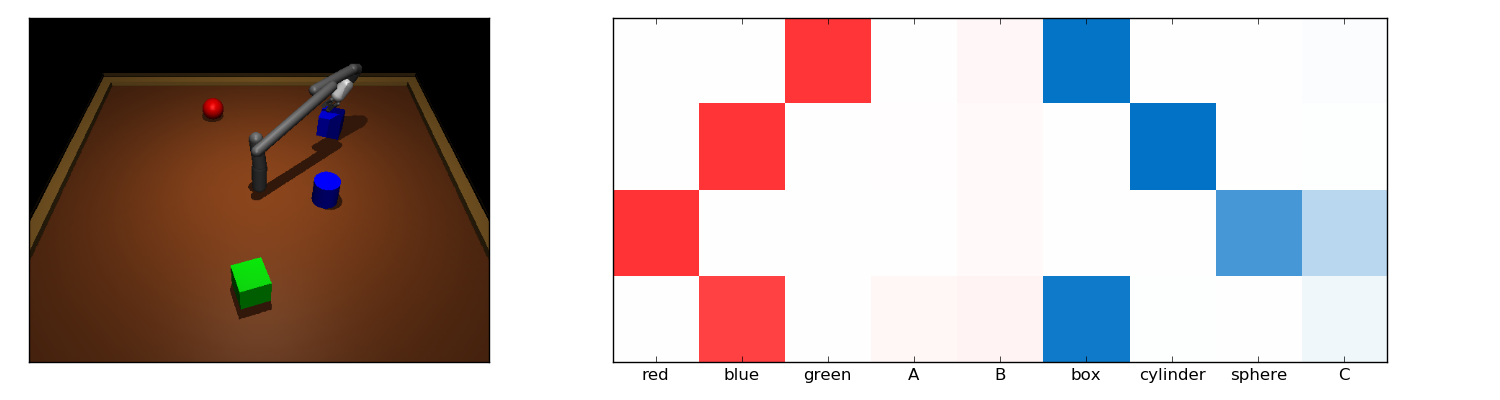}
\end{center}

Property identification is not always perfect.

\begin{center}
\includegraphics[width=0.92\linewidth]{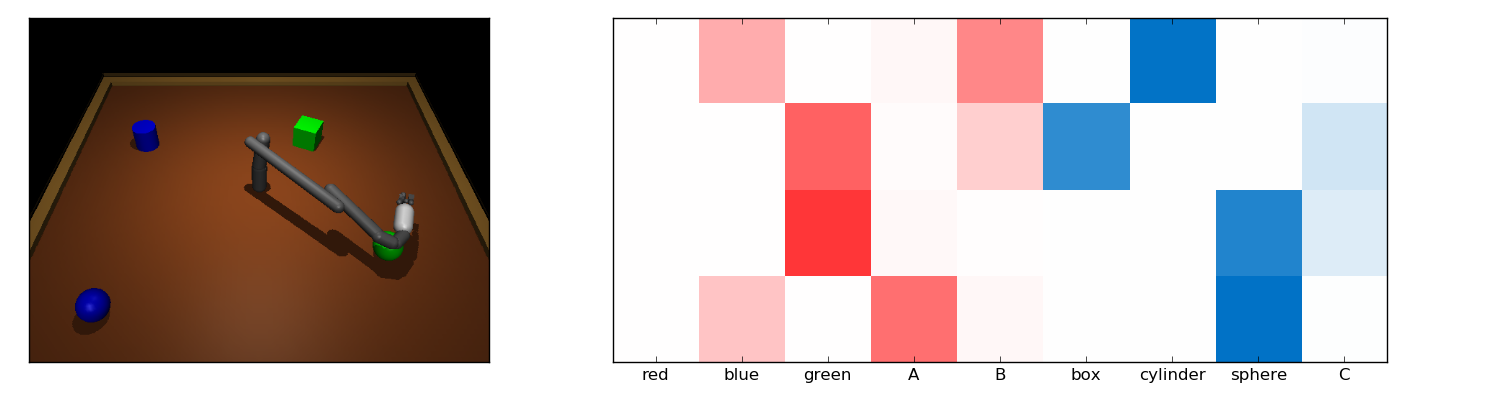}
\includegraphics[width=0.92\linewidth]{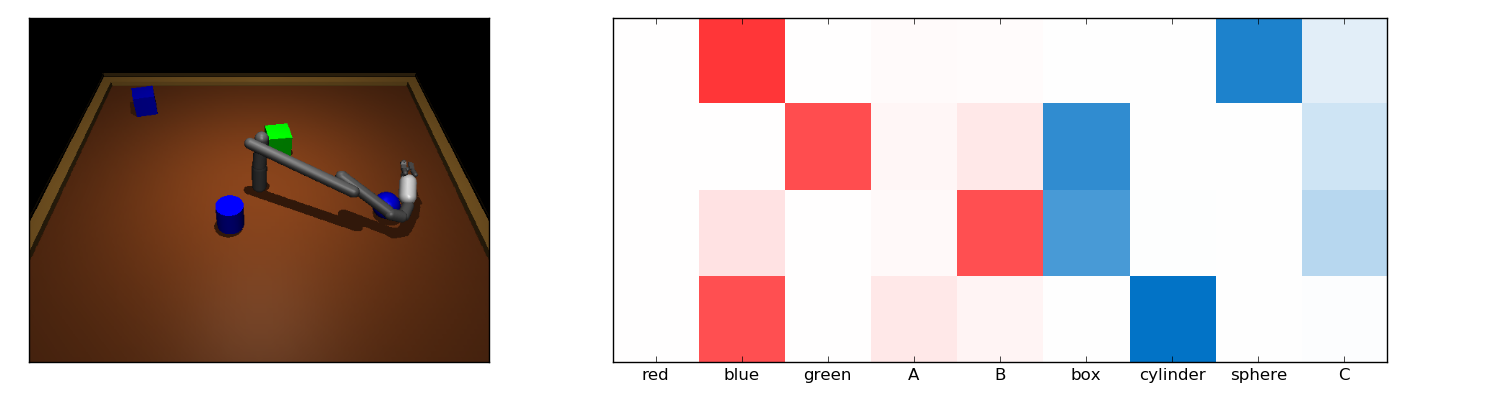}
\end{center}

\subsection{Behavior with previously unseen properties}

The agent has never seen capsules, or any magenta object.

\begin{center}
\includegraphics[width=0.92\linewidth]{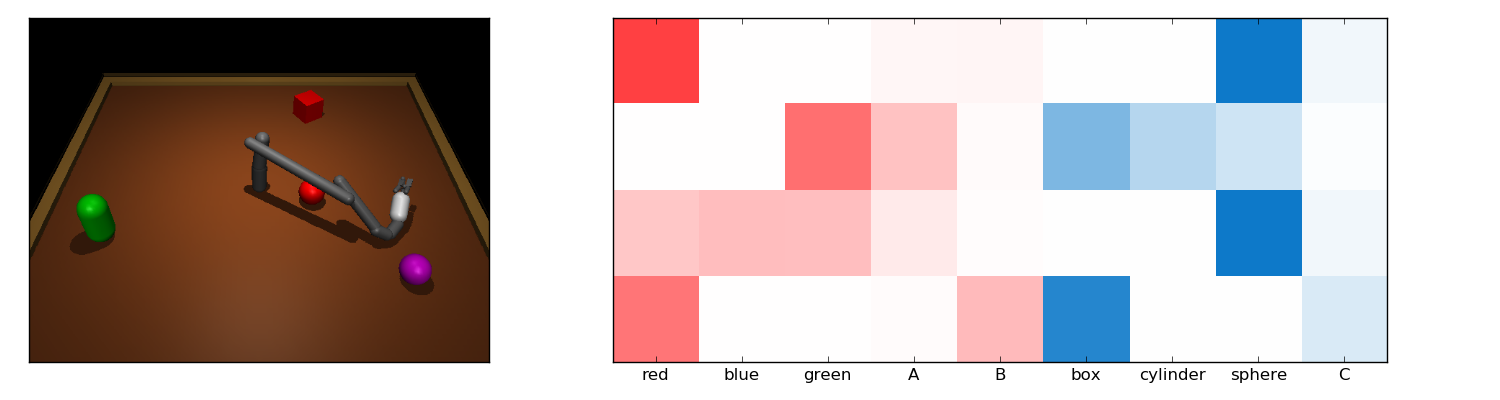}
\includegraphics[width=0.92\linewidth]{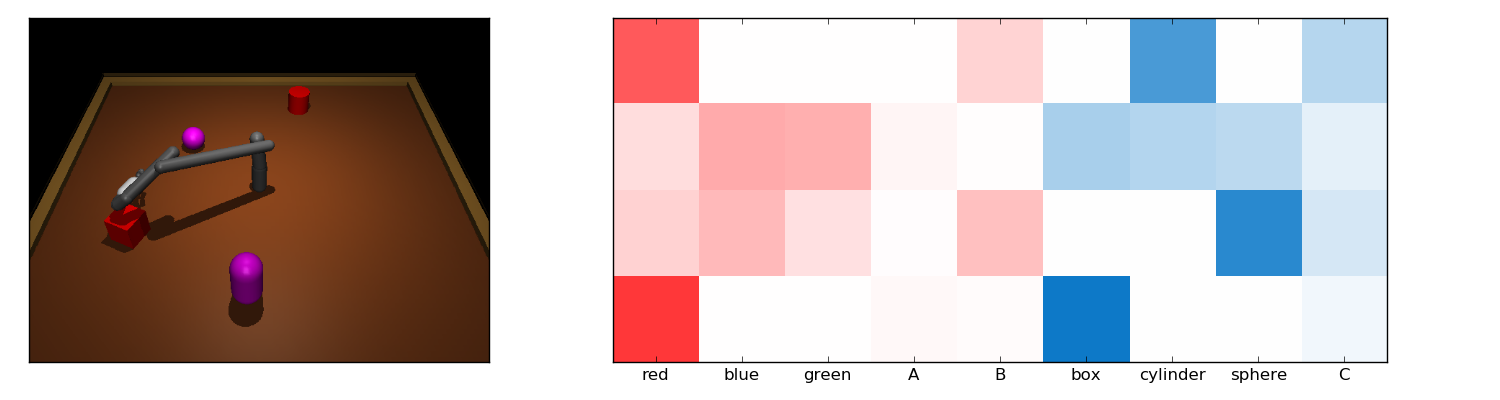}
\end{center}